# Can AI mimic the human ability to define neologisms?

**Georgios P. Georgiou**[1][2]

[1]Department of Languages and Literature, University of Nicosia, Nicosia, Cyprus

[2]Director of the Phonetic Lab

`georgiou.georg@unic.ac.cy`

**Abstract**

One ongoing debate in linguistics is whether Artificial Intelligence (AI) can effectively mimic human performance in language-related tasks. While much research has focused on various linguistic abilities of AI, little attention has been given to how it defines neologisms formed through different word formation processes. This study addresses this gap by examining the degree of agreement between human and AI-generated responses in defining three types of Greek neologisms: blends, compounds, and derivatives. The study employed an online experiment in which human participants selected the most appropriate definitions for neologisms, while ChatGPT received identical prompts. The results revealed fair agreement between human and AI responses for blends and derivatives but no agreement for compounds. However, when considering the majority response among humans, agreement with AI was high for blends and derivatives. These findings highlight the complexity of human language and the challenges AI still faces in capturing its nuances. In particular, they suggest a need for integrating more advanced semantic networks and contextual learning mechanisms into AI models to improve their interpretation of complex word formations, especially compounds.

*Keywords:* AI; blends; ChatGPT; compounds; derivatives; neologisms

## 1. Introduction

An ongoing and intriguing debate focuses on whether Large Language Models (LLMs) can replicate human language. The literature presents mixed evidence on this matter. Several studies suggest that LLMs can generate text closely resembling human language (Bubeck et al., 2023; Clark et al., 2021; Georgiou, 2025). However, the widely accepted concept of a universal grammar inherent in humans (Chomsky, 2000) challenges the idea that machine cognition can mirror human cognition. According to Chomsky et al. (2023), models like ChatGPT function as statistical engines driven by pattern recognition. Supporting this perspective, other studies highlight significant differences between human cognition and LLMs, which are reflected in language (Cai et al., 2024; Georgiou, 2024; Herbold et al., 2023). For instance, Georgiou (2024) examined how various linguistic components are represented in human-written and AI-generated texts, assessing the ability of ChatGPT to emulate human writing. The author found that despite AI-generated texts appearing to mimic human language, the results revealed significant differences across multiple linguistic features in the domains of phonology, grammar, and semantics.

The current body of research has primarily explored a narrow portion of the wide range of phenomena that define human language, concentrating mostly on syntax and semantics (Weissweiler et al., 2023). One area that has been notably overlooked is morphology, which refers to the ability to generate words based on systematic patterns of form and meaning variation (Haspelmath & Sims, 2013). This study focuses on the underexplored phenomenon of neologisms, aiming to assess the ability of an LLM to assign human-like definitions to different types of neologisms. These neologisms are derived from Greek, a language known for its rich morphological structure.

### 1.1 Definition of neologisms and work in Greek

The word *neologism* is of Greek origin and combines the words "néos" ("new) and "lógos" ("speech"). The Utilitarian Dictionary of Modern Greek (Academy of Athens, 2023) defines neologism as "the result of neology, a lexical unit introduced into the vocabulary of a language". According to Rodríguez Guerra (2016), the definitions of neologisms can be categorized into three main types: a) the general definition, which refers to new words or expressions, whether they are formally or semantically novel, and includes both internal processes (like derivation, composition) and external ones (such as loanwords or semantic calques); b) the definition based on a lexicographic criterion, which focuses on forms or expressions not yet recorded in general dictionaries, often originating from existing words or borrowed from other languages; and c) the definition based on usage over time criterion, which defines neologisms as words, phrases, or meanings that have gained popularity or vogue over a specific period. The author argues that although numerous definitions of neologism have been proposed, along with various lists outlining different criteria for identifying neologisms in a given language, these definitions consistently center on the concept of "newness" when referring to the lexicon of a language.

Research on Greek neologisms remains relatively underexplored. Most existing studies focus on various aspects, including the developmental and typological characteristics of neologisms (Anastasiadi-Symeonidi, 1986), their use in different historical periods of Greece (Eleutheriadou, 2018), the emergence of new terms during the COVID-19 pandemic (Stathopoulou, 2022), their translation (Lazarou, 2018), categorization (Michalopoulou, 2022), and their occurrence and usage in lexical databases and corpora (Sfinias, 2024), among others. Future work needs to align with current trends and investigate, for example, the processing of Greek neologisms by both humans and machines; such a venture is attempted through this study.

### 1.2 Word formation processes in Greek

This study investigates Greek neology blends, compounds, and derivatives, driven by their distinct morphological properties, which provide valuable insights into language processing. As a morphologically rich language, Greek serves as an ideal testing ground for examining how different word formation processes influence the definition of neologisms.

*Blending* is a word formation process where two or more words are combined into one, with the constituent parts either being shortened or overlapping partially (Beliaeva, 2019). Ralli and Xydopoulos (2011) inclined to propose that blending is an emerging process in Greek mostly found in subvarieties like youth language and other marginal varieties. For instance, the word "vlaks-itzís" ("stupid taxi driver") might be used as a slang term to denote irony. The authors stated that blends are likely to become more productive over time and applicable in general neologism formation, similar to its role in English. *Compounding* involves merging two words to create a single concept, where each word retains its individual meaning when considered separately, but together they represent a new, unified idea (Tenizi & Georgiou, 2020). Greek compound formation is part of the grammatical aspect of morphology, alongside derivation and, to a certain degree, inflection. Greek compounds include a semantically empty element, which is realized as /o/, and which it is placed between the two constituents like in the following example: "psar-o-tavérna" ("fish tavern"). However, there are cases in which /o/ is deleted when the second constituent starts with a weaker vowel: "lað-é-mboros" ("oil merchant") (Ralli, 2012). Another word formation process, which results in new words is *derivation* (Beard, 1998). Derivation in Greek is characterized by two main processes: the affixational process, which leads to the creation of new lexemes through the use of affixes (e.g., "anixnévo", "detect" → "anixnef-tís", "detector"), and the non-concatenative process, which leads to the generation of new lexemes, either with or without altering the form of the base to which they are applied (e.g., "psixr-ós", "cool" → "psíxr-a", "coolness"). In Standard Modern Greek, affixation is productive (Efthymiou, 2023); see some examples of prefixation and suffixation from Koutsoukos & Efthymiou (2023), e.g., "ðiacini-tís" ("trafficker"), "iper-aliévo" ("overfish").

Blends may be compared to compounds, as both processes include the unification of two or more lexemes and the creation of a single word with another meaning. In addition, they both display only one stress, combine the same grammatical categories, and are subject to form reduction (Ralli & Xydopoulos, 2011). However, Ronneberger-Sibold (2006) argued that the formation of blends arises from deliberate thought, whereas compounds are created effortlessly through established word formation processes. Ralli and Xydopoulos (2011) stated that blends are created by the speakers to achieve a particular effect in specific contexts with a clearly extragrammatical motivation, a purpose not served by standard compounds; this may include the expression of irony, the creation of a sense of mystery, conveyance of a playful tone, or the delivery an allusive message. Moreover, compounds are also different from derivatives in that the former combine items with a lexical content, while the latter encompass various affixes (Ralli, 2012).

### 1.3 AI and neologisms

Research on how neologisms are processed by AI systems is scarce. A recent study by Marelli and Amenta (2023) investigated the ability of ChatGPT to process and interpret novel derived words. While human speakers readily infer meanings of newly encountered words through morphological structure and semantic plausibility, computational models often struggle with low-frequency or novel items. This research examined whether ChatGPT could accurately define novel words and estimate their perceived meaningfulness as judged by human speakers. A dataset of English novel derived was used, each embedding different affixes. Human judgments of meaningfulness, collected in prior studies, served as a benchmark and ChatGPT was prompted to provide meaningfulness ratings on a 1–7 scale. These predictions were compared against human judgments and FRACSS, a cognitively inspired distributional semantics model that predicts word meaning based on morphological transformations. The quantitative analysis revealed that ChatGPT's meaningfulness ratings correlated with human judgments ($r = .25$, $p < 0.001$) and significantly improved the baseline prediction model. Further mixed-effects modeling demonstrated that the predictions of the model were highly dependent on affix frequency and productivity. Therefore, ChatGPT demonstrates competence in defining novel words but falls short in capturing human intuitions about their meaningfulness. Its reliance on affix-based heuristics, rather than full semantic integration, limits its effectiveness in modeling human word processing. This study presents novel words as a valuable challenge for LLMs and emphasizes the need for further research into their cognitive plausibility and underlying linguistic strategies.

Other studies that have explored the processing of various word formation processes also highlight challenges found in LLMs. A study analyzing the representations of compounds in AI models revealed only moderate alignment with human semantic intuitions (Buijtelaar & Pezzelle, 2023). This suggests that, unlike humans, AI struggles to integrate the meanings of individual components into a cohesive whole. Research from the University of Alberta demonstrated that when encountering words like "carpet", individuals instinctively parse them into "car" and "pet", despite the combined word having an unrelated meaning (Chamberlain et al., 2020). This process highlights our reliance on world knowledge and context to derive meaning, which is absent from LLMs. Similarly, AI models have been shown to process blends by focusing on the individual components, often without fully accessing the combined meaning. This limitation arises because the blending process can obscure the original components, making it challenging for AI to recover their meanings (Pinter et al., 2020). Nevertheless, there is evidence indicating good performance of AI models in morphological tasks. For example, Manova (2023) discusses how AI models like ChatGPT utilize subword units to process language. This approach allows AI systems to handle derivational morphology effectively by breaking down words into smaller, manageable units. A final conclusion of the author is that the processing of ChatGPT is not that different from human processing when it comes to morphology. Weller-Di Marco and Fraser (2023) examined the ability of ChatGPT to understand morphologically complex words. More specifically, the study focused on tasks like identifying the main noun in compound words, recognizing shared verb stems, or spotting incorrectly formed words. The findings showed that the language model performed well on most tasks, but struggled with identifying words that were formed incorrectly. While the model generally understood the structure of complex words, it did not seem to have formal knowledge of derivational rules. Instead, it relied on interpreting the word parts it had observed to figure out the meaning of the word.

**1.4 This study**

This study investigates the degree of agreement between human and AI-generated responses in defining three types of Greek neologisms: blends, compounds, and derivatives. The methodology involved an online experiment in which human participants selected the most appropriate definitions for neologisms, with the same prompts given to ChatGPT. The results were analyzed using descriptive statistics and statistical comparisons. To the best of the researcher's knowledge, such investigations are absent from the existing literature and can provide valuable insights into how AI processes novel word formations in comparison to human cognition. By addressing this gap, the study enhances our understanding of the morphological capabilities and limitations of AI, offering a foundation for future research aimed at refining AI-driven language models. It is hypothesized that human responses will overlap only to some extent with those of AI, due to current limitations of AI models in handling neologisms based on previous findings. Compounds and blends are expected to overlap the least due to the difficulty AI experiences in extracting the meaning of words formed by merging other words, while derivatives might exhibit a higher degree of overlap.

## 2. Methodology

### 2.1 Participants

The participants of the study were 30 speakers ($n_{males}$ = 2) in the age range of 23–48 ($M$age = 31.8, $SD$ = 6.33). They were either current or former MA students of Greek. All participants were native speakers of Standard Modern Greek with at least B2-level proficiency in English; in addition, they possessed knowledge of other languages, including French, German, Italian, Spanish, and others at varying levels. The participants had never experienced any language or cognitive disorders.

### 2.2 Instrument

An online survey was created in SoSci (Leiner, 2021). The first part of the survey required participants to provide demographic information such as age, first language, foreign languages and their level according to the Common European Framework for Languages, and gender. The second part included a familiarization test with three words, followed by the main test, which included 30 words. There were three types of words: blends, compounds, and derivatives. The words were nonexisting and created by the author. More specifically, blended noun words were constructed by combining parts of two real words. Noun compounds were formed by joining two existing words, and noun derivatives involved adding affixes to root words. The neologisms adhered to the structure and characteristics of real Greek words by following phonotactics, stress patterns, and morphological rules of Greek. The words were also tested with native Greek speakers to ensure they felt Greek. Furthermore, the three definitions of each neologism were created using feedback from five Greek speakers, who were presented with the target words and then asked to give possible definitions. The author selected the three more frequent definitions for each word and included them in the survey.

### 2.3 Procedure

The survey was administered online and in Greek. It started by asking participants to complete their demographics. In the main test, participants were presented with invented words and asked to select one of three potential definitions that they believed best corresponded to the word. They were informed that there were no correct or incorrect choices, as the words were entirely fabricated, and no contextual information was provided to suggest what might be more or less meaningful. The focus of the study was on participants' judgments regarding what interpretation seemed most suitable to them when encountering the words in isolation. Moreover, participants were prompted to decide quickly and spontaneously. To ensure that they understood the requirements of the survey, a small pilot test with three items was completed before the main test. The average completion time for the survey was 10.93 min ($SD$ = 3.65). All data collected remained completely anonymous and participants were informed about their rights and all ethical aspects at the beginning of the survey.

## 3. Results

For the investigation of the degree of agreement between the responses of humans and AI, we used Cohen's kappa; it comprises a statistical measure of interrater reliability or agreement between two raters, accounting for the possibility of agreement occurring by chance. Unlike a simple percent agreement, which only reflects the proportion of identical classifications, kappa evaluates the observed agreement relative to what would be expected if the raters were assigning classifications randomly. Kappa values range from –1 (perfect disagreement) to 1 (perfect agreement). According to Cohen, kappa values can be interpreted as follows: ≤ 0: no agreement; 0.01–0.20: slight or no agreement; 0.21–0.40: fair agreement; 0.41–0.60: moderate agreement; 0.61–0.80: substantial agreement; and 0.81–1.00: almost perfect agreement (McHugh, 2012). We have also employed the overlap score, which quantifies the agreement between two datasets, specifically between responses from the human participants and the AI. This score is computed using the formula: overlap score = number of matching categories / total number of categories. The categories represented each of the three responses which correspond to a different definition of a neologism. The overlap score is expressed as a proportion.

All statistical analyses were conducted in R (R Core Team, 2025). First, we calculated the degree of agreement in the responses across participants for each type of neologism. The Fleiss' Kappa value indicated a slight degree of agreement among participants for the responses of blends ($\kappa = 0.11$; $p < 0.001$), compounds ($\kappa = 0.11$; $p < 0.001$), and derivatives ($\kappa = 0.16$; $p < 0.001$); the results indicated that participants showed minimal consistency in their responses. Regarding the human and AI responses, the results demonstrated fair agreement on average based on the kappa values for blends and derivatives, whereas compounds yielded no agreement. In addition, the results revealed moderate overlap between the responses of humans and AI for blends and derivatives, while lower overlap was observed for compounds. Table 1 presents the degree of agreement between each human and the AI in their responses based on the kappa values and the overlap scores. Figures 1 and 2 illustrate the mean kappa values and overlap scores between human and AI responses for each type of neologism.

Table 1: Degree of agreement between responses from each human and the AI responses for each type of neologism based on kappa values and overlap scores

| Participant | Blend Kappa (p–value) | Overlap score | Compound Kappa (p–value) | Overlap score | Derivative Kappa (p–value) | Overlap score |
|---|---|---|---|---|---|---|
| 1 | 0.26 (0.18) | 0.50 | 0.43 (0.03) | 0.60 | –0.38 (0.08) | 0.10 |
| 2 | 0.17 (0.35) | 0.40 | 0.06 (0.77) | 0.40 | 0.56 (0.01) | 0.70 |
| 3 | 0.06 (0.77) | 0.40 | –0.25 (0.15) | 0.10 | 0.53 (0.01) | 0.70 |
| 4 | 0.22 (0.31) | 0.50 | 0.09 (0.67) | 0.40 | 0.41 (0.05) | 0.60 |
| 5 | 0.17 (0.46) | 0.50 | 0.00 (1.00) | 0.40 | 0.23 (0.30) | 0.50 |
| 6 | –0.11 (0.54) | 0.20 | –0.08 (0.58) | 0.20 | 0.39 (0.08) | 0.60 |
| 7 | 0.26 (0.18) | 0.50 | –0.18 (0.39) | 0.20 | 0.54 (0.02) | 0.70 |
| 8 | 0.41 (0.03) | 0.60 | –0.03 (0.89) | 0.30 | 0.23 (0.28) | 0.50 |
| 9 | –0.17 (0.46) | 0.30 | –0.25 (0.24) | 0.20 | –0.25 (0.24) | 0.20 |
| 10 | 0.53 (0.01) | 0.70 | 0.29 (0.14) | 0.50 | 0.40 (0.06) | 0.60 |
| 11 | 0.31 (0.08) | 0.50 | –0.06 (0.78) | 0.30 | 0.25 (0.25) | 0.50 |
| 12 | 0.06 (0.77) | 0.40 | –0.06 (0.78) | 0.30 | 0.55 (0.02) | 0.70 |
| 13 | 0.06 (0.77) | 0.40 | 0.29 (0.14) | 0.50 | 0.08 (0.73) | 0.40 |
| 14 | 0.06 (0.77) | 0.40 | 0.09 (0.67) | 0.40 | 0.26 (0.20) | 0.50 |
| 15 | 0.41 (0.03) | 0.60 | 0.12 (0.58) | 0.40 | 0.55 (0.01) | 0.70 |
| 16 | 0.26 (0.17) | 0.50 | 0.24 (0.14) | 0.50 | 0.25 (0.25) | 0.50 |
| 17 | 0.22 (0.31) | 0.50 | –0.06 (0.76) | 0.30 | 0.41 (0.05) | 0.60 |
| 18 | 0.22 (0.26) | 0.50 | –0.36 (0.06) | 0.10 | 0.05 (0.81) | 0.40 |
| 19 | 0.82 (0.00) | 0.90 | –0.17 (0.47) | 0.30 | –0.06 (0.78) | 0.30 |
| 20 | 0.21 (0.17) | 0.40 | –0.18 (0.30) | 0.20 | 0.38 (0.08) | 0.60 |
| 21 | 0.26 (0.17) | 0.50 | –0.03 (0.88) | 0.30 | 0.06 (0.77) | 0.40 |
| 22 | 0.38 (0.08) | 0.60 | 0.56 (0.01) | 0.70 | 0.54 (0.02) | 0.70 |
| 23 | 0.46 (0.05) | 0.70 | –0.13 (0.58) | 0.30 | 0.25 (0.23) | 0.50 |
| 24 | 0.41 (0.04) | 0.60 | –0.06 (0.78) | 0.30 | 0.69 (0.00) | 0.80 |
| 25 | 0.04 (0.87) | 0.50 | –0.18 (0.39) | 0.20 | 0.25 (0.23) | 0.50 |
| 26 | 0.50 (0.03) | 0.70 | –0.03 (0.89) | 0.30 | 0.41 (0.05) | 0.60 |
| 27 | 0.50 (0.02) | 0.70 | 0.19 (0.39) | 0.50 | 0.41 (0.05) | 0.60 |
| 28 | –0.07 (0.76) | 0.40 | 0.22 (0.25) | 0.50 | 0.08 (0.73) | 0.40 |
| 29 | 0.83 (0.00) | 0.90 | –0.09 (0.53) | 0.30 | 0.37 (0.07) | 0.60 |
| 30 | 0.21 (0.17) | 0.40 | –0.21 (0.28) | 0.20 | 0.22 (0.30) | 0.50 |
| Average | 0.27 | 0.52 | 0.01 | 0.34 | 0.29 | 0.53 |
| (SD) | (0.24) | (0.16) | (0.21) | (0.14) | (0.24) | (0.16) |

Two one-way ANOVA tests were applied to investigate potential differences in kappa and overlap scores among the different types of neologisms. The tests manifested significant effects of type on the kappa value [$F(2,87) = 13.82$, $p < 0.001$] and the overlap score respectively [$F(2,87) = 15.37$, $p < 0.001$]. Subsequent posthoc tests with the Tukey method revealed significant differences between blend and compound (kappa: $\beta = 0.26$, $SE = 0.06$, $t = 4.34$, $p < 0.001$; overlap scores: $\beta = 0.13$, $SE = 0.04$, $t = 3.6$, $p < 0.001$), and compound and derivative (kappa: $\beta = –0.28$, $SE = 0.05$, $t = –4.74$, $p < 0.001$; overlap score: $\beta = -$

0.19, $SE = 0.04$, $t = -4.92$, $p < 0.001$); no differences occurred between blend and derivative (kappa: $β = –0.24$, $SE = 0.05$, $t = –0.39$, $p = 0.92$; overlap score: $β = –0.01$, $SE = 0.04$, $t = –0.25$, $p = 0.96$).

For further examination of the degree of agreement, we calculated the mode or majority category for each item and compared this aggregated result to the AI responses. It is observed that blends received the highest values followed by derivatives and compounds. Blends and derivatives exhibited almost perfect and substantial agreement respectively, as well as high overlap scores, while compounds exhibited no agreement and a low overlap score. Table 2 shows the degree of agreement between the human responses (majority category) and the AI responses for each type of neologism based on kappa values and overlap scores.

Table 2: Degree of agreement between the human responses (majority category) and the AI responses for each type of neologism based on kappa values and overlap scores.

| All Participants | Blend | | Compound | | Derivative | |
|---|---|---|---|---|---|---|
| | Kappa (*p*-value) | Overlap score | Kappa (*p*-value) | Overlap score | Kappa (*p*-value) | Overlap score |
| | 0.83 (< 0.001) | 0.90 | 0 (1) | 0.30 | 0.69 (0.001) | 0.80 |

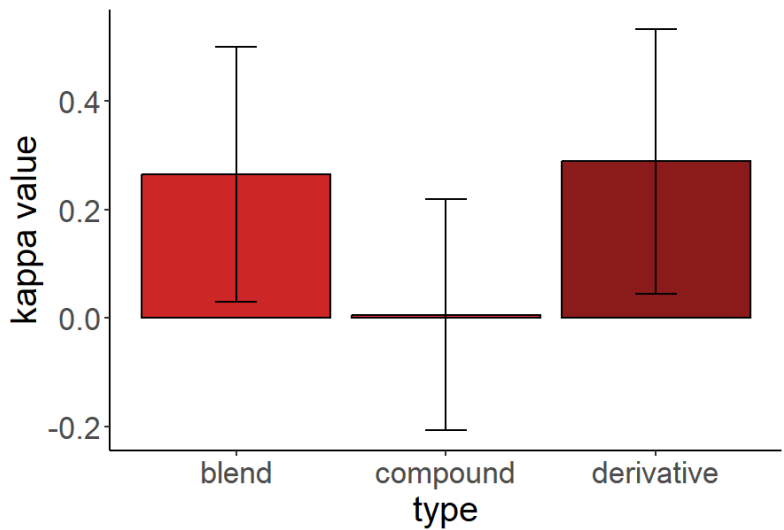

Figure 1: Mean kappa values between human and AI responses for each type of neologism.

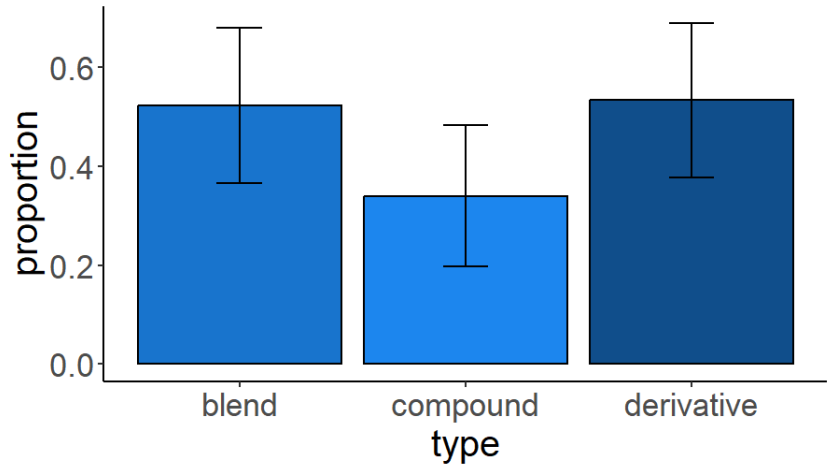

Figure 2: Mean overlap scores between human and AI responses for each type of neologism.

## 4. Discussion

The study investigated how various types of neologisms are defined by humans and AI. More specifically, human participants with a Greek native background were asked to choose the definition of invented Greek neologisms such as blends, compounds, and derivatives, from three alternative responses. The same procedure was conducted by ChatGPT. The responses of the two groups were compared using agreement scores.

The results showed that when averaged per participant, the kappa values and overlap scores for AI and human agreement are low but when comparing the mode/majority category for each item to AI responses, they are higher. The first scenario corroborated the variation observed in individual human responses, meaning that humans do not consistently align with AI at the personal level. Therefore, AI is not mimicking any single individual's decision-making process but may align more with a general trend. Concerning the second scenario, the high agreement suggests that the responses of AI are more in line with what the majority of humans select rather than what any one individual chooses. This implies that AI may be capturing the most common or dominant interpretation of the task rather than reflecting the variability found in human responses. So, if AI is intended to approximate the best guess or common agreement among humans, then the high mode-based agreement suggests that AI is effective for that purpose.

The statistical analysis indicated that blends and derivatives had a similar agreement between humans and AI, while compounds had a significantly lower degree of agreement. The observation that AI models often struggle to interpret compound words as humans do is supported by the fact that compounds often derive meaning from semantic and contextual relationships between their components, which might be limited in AI. Humans might rely more on world knowledge and context-driven inference, whereas AI may be limited to pattern recognition without deep semantic understanding (Chamberlain et al., 2020). LLMs generate text without truly understanding meaning or having a sense of the readers' thoughts. Even though the text might seem natural and fluent, it is not based on real communication or intent. The model does not have an actual thought process or a way to know what it is trying to say – it is just putting together words based on patterns it learned from data. When humans read an AI-generated text, they often think it has meaning because of the human ability to interpret language, but the model itself does not know what it is conveying (Bender et al., 2021).

The observed discrepancy in the definition of neologisms by AI – particularly its performance with derivatives (and blends) compared to compounds – suggests that AI systems may rely more on form-based cues than on context-driven semantic understanding; consequently, AI models may process derivatives more easily. This is supported by studies suggesting that AI models, including ChatGPT, are capable of processing derivational morphology effectively (e.g., Manova, 2023). Moreover, the greater agreement on derivatives compared to compounds may be attributed to the fact that affixes in Greek typically carry their own meaning, allowing the meaning of a word to be inferred from its affixes. For example, the Greek word "a-δίnamos" ("weak") would easily be understood by AI through morphological decomposition. Since Greek affixes typically have consistent meanings, AI can analyze the word by breaking it down into its components – a- ("without") – and then reconstructing the overall meaning; as a result, "a-δίnamos" is interpreted as "someone without strength", or "weak". This would align with the reliance of ChatGPT on the meanings of individual word components to interpret the whole word (Weller-Di Marco & Fraser, 2023). However, contrary to the expectations, blends did not yield lower overlap compared to derivatives. This is consistent with the limitations observed by Pinter (2020). Perhaps the Greek blends used in this study, even though they involve fusion, maintain a higher degree of semantic transparency than compounds. The compounds might be more semantically opaque, making more difficult the extraction of the original meanings of the components. The AI can reasonably infer the meanings of the blends' components based on its knowledge of these components. For example, in the word "texno-tomía," both humans and AI may have interpreted the first component, "texno-," as deriving from the frequently used word "texno-λογία" ("technology"), and the second component, "-tomía," as deriving from "keno-tomía" ("innovation"). As a result, they most likely chose the definition of "the process of developing *innovative* products using advanced *technologies*".

Nevertheless, it needs to be taken into account that despite the high degree of agreement between human and AI responses for blends and derivatives as a general trend, the morphological capabilities of humans exhibit only fair agreement with those of AI, as seen in the averaged participant responses. In support of this, previous work demonstrates that while ChatGPT presents impressive linguistic capabilities, its morphological understanding, particularly in handling novel word forms, remains limited and does not yet parallel human-like language skills (Weissweiler, 2023).

The findings could be relevant for second language acquisition and the further refinement of AI-based natural language processing models. Given the important benefits of generative AI tools such as ChatGPT for linguistic analysis by students, as highlighted in the literature (e.g., Loock, & Holt, 2024), the findings suggest that different types of word formation, such as blends, compounds, and derivatives, may present varying levels of difficulty for AI-based learning tools to teach or assess. This is particularly relevant in the context of language learning, where AI-assisted systems are increasingly used to support vocabulary acquisition. AI systems may struggle with effectively modeling and presenting compounds. To enhance the efficacy of AI-assisted language learning, these systems may require additional linguistic modeling and more sophisticated algorithms to handle specific word formation types. Such improvements could lead to more accurate vocabulary teaching and comprehension support, ultimately boosting the effectiveness of these tools in helping learners acquire and understand new words in their second language.

## 5. Conclusion

The present study revealed fair agreement between human definitions and the corresponding definitions selected by ChaptGPT for most types of Greek neologisms. This observation underscores the complexity of human language and the challenges that remain in developing AI systems that can fully capture its nuances. However, agreement was higher when the majority of human responses were considered, revealing the effectiveness of AI in capturing general trends. The study also yielded a higher overlap between human and AI responses for blends and derivatives versus compounds; this highlights the current limitations of AI in semantic processing. While AI excels at identifying and processing form-based cues in language, it struggles with interpreting meanings that rely on broader context or world knowledge. This suggests a need for integrating more sophisticated semantic networks and contextual learning mechanisms into AI language models to enhance their understanding of complex word formations. Furthermore, the focus of the study on Greek is important. Languages differ in their morphology and word formation processes. The findings may not be directly generalizable to all languages, calling for similar studies in other languages to understand the interplay between language-specific features and the ability of AI to interpret neologisms. Finally, future research could extend the comparison to multiple AI systems beyond ChatGPT to assess their ability to define neologisms. This would provide a more comprehensive understanding of their linguistic capabilities and help guide improvements in their morphological and semantic processing.

## Acknowledgments

This study is part of the COST Action – European Network On Lexical Innovation (ENEOLI) funded by the European Union (https://eneoli.eu/). I would like to thank the participants of the study and the reviewers for their valuable comments. Special thanks also to Dr Esther Breuer for leading Working Group 3 – Phonological, orthographic, grammatical, and semantic inclusion of neologisms in our mental lexicon, as well as the other group members for their valuable comments and ideas.

## Conflict of interests

The author declares no conflicts of interest.